\documentclass[11pt]{article}

\usepackage{acl}

\usepackage{times}
\usepackage{latexsym}
\usepackage[T1]{fontenc}
\usepackage[utf8]{inputenc}
\usepackage{microtype}
\usepackage{inconsolata}

\usepackage{amsmath}
\usepackage{graphicx}
\usepackage{hyperref}
\usepackage{algorithm}
\usepackage{algpseudocode}
\usepackage{multirow}
\usepackage{makecell}
\usepackage{diagbox}
\usepackage{siunitx}
\usepackage{booktabs} 
\usepackage{enumitem}
\usepackage{xcolor}

\usepackage[capitalize,noabbrev]{cleveref}

\title{ASKD-Whisper: Adaptive Self-knowledge Distillation for Efficient and Low-Latency Automatic Speech Recognition}

\author{
  \textbf{Junseok Lee\textsuperscript{1}},
  \textbf{Nahun Kim \textsuperscript{1}},
  \textbf{Chang-Jae Chun\textsuperscript{2}}
\\
\\
  \textsuperscript{1} OKESTRO Co., Ltd,
  \textsuperscript{2} Sejong University
}

\begin{document}
\maketitle

\begin{abstract}
Knowledge distillation (KD) is one of the most effective paradigms for compressing large-scale foundation models into deployable architectures. In the context of Automatic Speech Recognition (ASR), previous studies have predominantly focused on forcing the student model to strictly mimic the predictive distribution of a massive teacher model. However, this static dependency often presents an inherent trade-off: while the student rapidly acquires basic linguistic representations, it simultaneously inherits the teacher's domain-specific blind spots and over-confident hallucinations, leading to a severe decline in out-of-distribution generalization capacity. To effectively mitigate this issue, we propose Adaptive Self-Knowledge Distillation (ASKD), a dynamic curriculum framework. ASKD systematically decays the dependency on the teacher's distribution as training progresses—thereby unlocking the student's independent reasoning capacity—and subsequently employs a self-knowledge distillation phase to act as a structural regularizer. By applying ASKD, we distill the massive Whisper architecture into a compact variant, ASKD-Whisper. In our comprehensive evaluations across diverse acoustic domains, ASKD-Whisper not only achieves a 5$\times$ speedup in inference latency but also outperforms its teacher model by yielding a 1.07\% lower word error rate (WER). These results demonstrate that ASKD effectively prevents teacher-induced overfitting and establishes a new state-of-the-art for generalizable model compression.
\end{abstract}

\section{Introduction}
Automatic speech recognition (ASR) is becoming increasingly critical as its applications expand from basic transcription services to complex, real-time multimodal interactions involving human-to-robot communication, emergency response centers, and semantic search engines \cite{Feng_21, Haeb_21}. Driven by rapid advancements in deep learning (DL) and hardware acceleration, modern ASR architectures have shifted from traditional Hidden Markov Models to end-to-end neural frameworks \cite{Ren_19}. Recently, models leveraging transformer-based architectures and contrastive learning \cite{Baevski_20, Zhang_20} have achieved substantial performance improvements, effectively reaching human-parity in constrained environments. 

More recently, the paradigm has shifted toward massive foundation models. Whisper \cite{Radford_23}, trained on 680,000 hours of weakly supervised data, demonstrates remarkable zero-shot generalization across diverse domains and languages. Following the empirical scaling laws, researchers continually push performance boundaries by scaling up both model parameters and dataset sizes \cite{Ben_22}. However, deploying these billion-parameter models in real-world, latency-sensitive applications introduces severe system-level bottlenecks, rendering them impractical for edge devices or high-throughput real-time APIs \cite{Li_21}.

To deploy large-scale ASR models in practical systems, various compression techniques have been extensively explored \cite{Dudziak_19, Mehrotra_20, Rathod_22}. Among these, knowledge distillation (KD) \cite{Hinton_15} has been widely adopted across natural language processing (NLP), computer vision (CV), and ASR \cite{Gou_21}. The core premise of KD is to transfer the "dark knowledge" encoded in the soft probability distributions of a massive teacher model to a lightweight student model. In sequence-to-sequence ASR, researchers have minimized the Kullback–Leibler (KL) divergence of teacher distributions \cite{Gandhi_23} and aligned hidden features via masked token similarity transfer \cite{Choi_23}. Additionally, massive pseudo-labeling (PL) has been utilized to provide sequence-level distillation signals \cite{Gandhi_23}.

Despite these efforts, standard KD processes often fail to improve—and sometimes even degrade—the generalization capacity of the student model \cite{Stanton_21}. We identify a critical limitation in existing literature: static distillation forces the student to perfectly replicate the teacher throughout the entire training cycle. This over-supervision leads to a phenomenon where the student overfits to the teacher's specific inductive biases, inheriting its over-confidence on noisy data and linguistic hallucinations on ambiguous audio segments. 

To break this generalization bottleneck, we propose a novel dynamic framework: \textbf{Adaptive Self-Knowledge Distillation (ASKD)}. Our contributions are threefold:
\begin{itemize}[leftmargin=*]
    \item We introduce ASKD, a curriculum-based distillation paradigm that dynamically decouples the student from the teacher. By gradually transitioning from teacher-guided learning (AKD) to regularized self-training (SKD), we prevent teacher-induced overfitting and significantly enhance out-of-domain generalization.
    \item We present ASKD-Whisper, a highly optimized ASR model trained via ASKD. By retaining the powerful pre-trained Whisper encoder \cite{Radford_23} while utilizing a custom lightweight decoder, ASKD-Whisper achieves optimal speech-language alignment with a fraction of the computational cost.
    \item Through rigorous evaluation on diverse, real-world datasets (including noisy meetings and financial contexts), we demonstrate that ASKD-Whisper not only achieves a 5$\times$ inference speedup but outperforms its much larger teacher model, validating the superiority of the ASKD framework.
\end{itemize}

\section{Related Work}

\subsection{Large-Scale Speech Foundation Models}
The landscape of ASR has been reshaped by the introduction of foundation models. Whisper \cite{Radford_23} proved that massive scaling of weakly supervised data could yield highly robust representations without the need for meticulous human annotation. Subsequent works like Canary \cite{Krishna_24} and Qwen-Audio have extended these capabilities toward multilingual and multimodal understanding. While these models offer state-of-the-art accuracy, their autoregressive decoding complexity grows quadratically, making real-time streaming or low-latency processing nearly impossible without high-end GPU clusters. This underscores the urgent need for structural compression techniques that preserve the rich semantic representations of these foundation models.

\subsection{Knowledge Distillation in Seq2Seq Models}
Standard KD \cite{Hinton_15} relies on matching the output logits of a teacher and student. In sequence-to-sequence models, however, the temporal alignment of tokens makes distillation notoriously difficult. Distil-Whisper \cite{Gandhi_23} successfully addressed this by employing pseudo-labeling and KL divergence over a curated dataset. However, recent NLP research \cite{Stanton_21, Su_23} has highlighted the "capacity gap" problem: a small student model simply cannot absorb the full complexity of a massive teacher. Forcing it to do so using a static distillation weight ($\alpha$) often results in the student memorizing the teacher's prediction errors instead of learning generalizable linguistic rules. Self-Knowledge Distillation (SKD) \cite{Kim_21} emerged as an alternative, allowing a model to distill knowledge from its own previous epochs to act as a label-smoothing regularizer. Our proposed ASKD bridges these two paradigms, leveraging external teacher knowledge for rapid initialization while utilizing internal self-knowledge for ultimate generalization.

\section{Methodology}
\begin{figure*}[!ht]
    \centering
    \includegraphics[width=1.7\columnwidth]{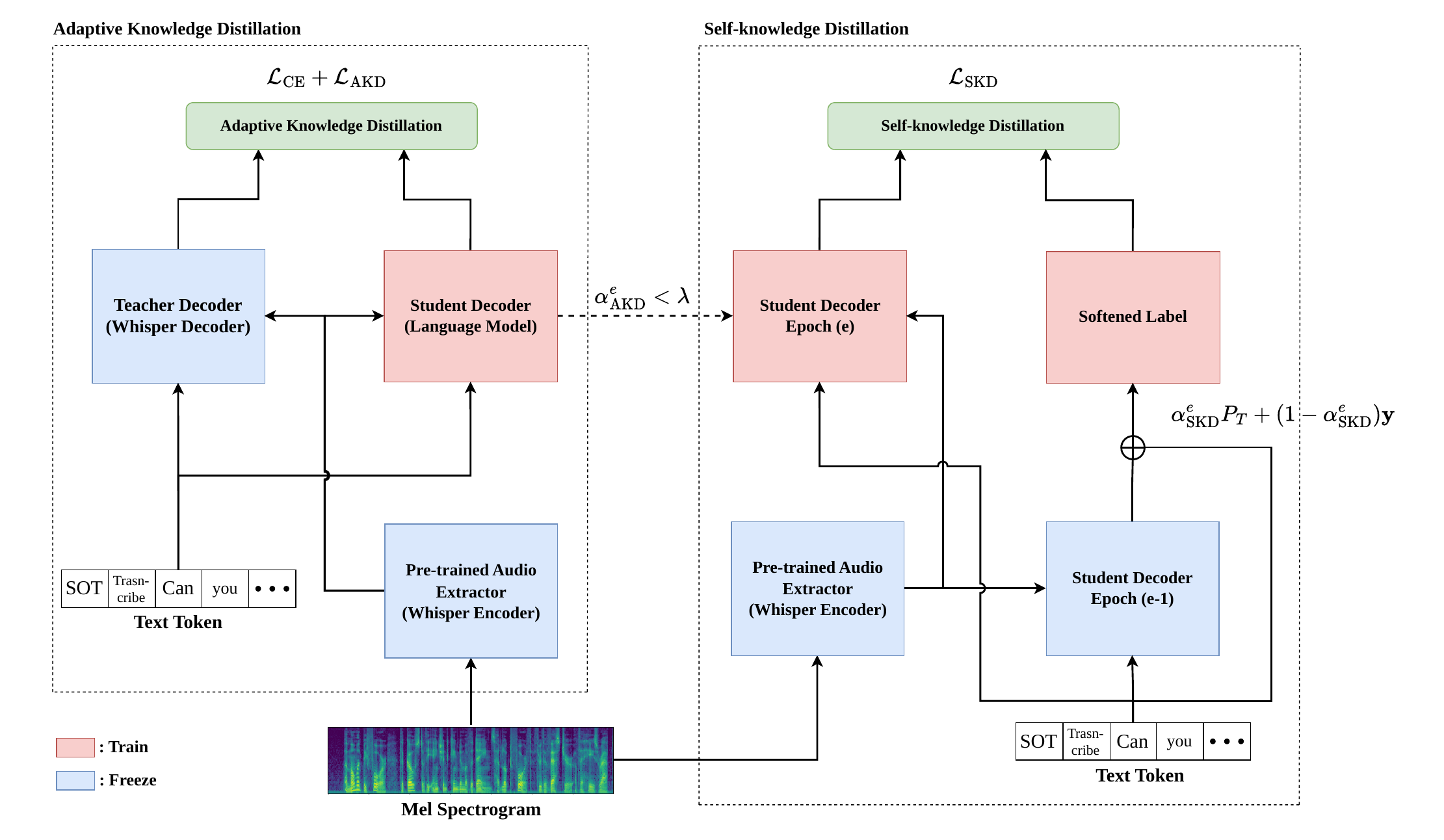}
    \caption{The overall architecture of the Adaptive Self-Knowledge Distillation (ASKD) framework. During the early epochs, AKD dynamically decays the distillation weight ($\alpha_\text{AKD}^e$) to prevent teacher-induced overfitting. Once $\alpha_\text{AKD}^e$ crosses the threshold $\lambda$, the system transitions to SKD, acting as a structural regularizer to enhance independent generalization.}
    \label{fig:flow-chart}
\end{figure*}

The overall architecture of the proposed ASKD framework is illustrated in Fig. \ref{fig:flow-chart}. The training process is bifurcated into two distinct temporal phases. First, the student's foundational representation capacity is rapidly established through Adaptive Knowledge Distillation (AKD). Second, once a critical capacity threshold is reached, the framework shifts to Self-Knowledge Distillation (SKD) to refine linguistic generalization independently of the teacher.

\subsection{Phase 1: Adaptive Knowledge Distillation}
Standard KD statically assigns $\alpha_{\text{KD}}$, a constant weight that determines the magnitude of the distillation penalty throughout the entire training process. However, enforcing a high static $\alpha_{\text{KD}}$ forces the student to blindly mimic the teacher, severely crippling the student's ability to learn generalized grammatical or acoustic patterns from the raw hard labels \cite{Su_23}. Conversely, starting with a low $\alpha_{\text{KD}}$ starves the student of necessary semantic guidance during the crucial early stages of training.

To resolve this dilemma, AKD treats the distillation weight as a dynamic curriculum. We initialize training with a high $\alpha_{\text{AKD}}^{initial}$ to rapidly transfer the teacher's robust vocabulary and acoustic mappings. After a predefined warm-up period, we monotonically decay this dependency. The AKD loss, $\mathcal{L}_\text{AKD}$, is formulated as:
\begin{align}
\label{eq:AKD_loss}
    \mathcal{L}_{\text{AKD}} = \alpha_{\text{AKD}}^{e} \tau^2 \mathcal{L}_{\text{KL}}(P_S^\tau || P_T^\tau)
\end{align}
where $e$ denotes the epoch index, $\mathcal{L}_{\text{KL}}$ represents the Kullback-Leibler divergence, and $P_S^\tau, P_T^\tau$ are the predictive distributions scaled by temperature $\tau$.
The decay function for $\alpha_{\text{AKD}}^{e}$ is defined as:
\begin{align}
\label{eq:AKD_update}
    \alpha_{\text{AKD}}^e = \alpha_{\text{AKD}}^{initial} - \frac{e-E_w}{E_t}
\end{align}
where $E_t$ is the total number of epochs and $E_w$ denotes the warm-up epochs. By gradually reducing $\mathcal{L}_\text{AKD}$, the student is forced to rely more heavily on the standard Cross-Entropy loss from the ground-truth data. This explicitly prevents the student from memorizing the teacher's token-level hallucinations, leading to a more robust internal representation \cite{Ganguly_24}.

\subsection{Phase 2: Adaptive Self-Knowledge Distillation}
Once $\alpha_{\text{AKD}}^{e}$ decays below a critical threshold $\lambda$, the student has acquired sufficient foundational knowledge. At this juncture, continuing to distill from a rigid external teacher yields diminishing returns. Thus, ASKD seamlessly transitions into the SKD phase.

In SKD, we generate soft labels using the student's own predictive distribution from the previous epoch ($P_S^{(e-1)}$), allowing the model to act as its own teacher. The SKD loss, $\mathcal{L}_\text{SKD}$, is calculated by computing the Cross-Entropy over a dynamically softened target:
\begin{align}
\label{eq:SKD_loss}
    \mathcal{L}_{\text{SKD}} = \mathcal{L}_{\text{CE}}\left((1 - \alpha_{\text{SKD}}^{e})\mathbf{y} + \alpha_{\text{SKD}}^{e} P_S^{(e-1)}, P_S^{(e)}\right)
\end{align}
where $\mathbf{y}$ is the one-hot hard label. Over-supervision on hard labels $\mathbf{y}$ typically leads to severe overfitting, as the model becomes pathologically confident in noisy acoustic environments \cite{Kim_21}. By injecting its own softened historical predictions, the student learns the inter-class probabilities and semantic relationships between phonetically similar tokens. 

Crucially, because the student's early representations were stabilized by the teacher during Phase 1, the historical predictions $P_S^{(e-1)}$ are highly reliable. As training progresses, we increase the reliance on these soft labels:
\begin{align}
\label{eq:SKD_update}
    \alpha_{\text{SKD}}^e = \alpha_{\text{SKD}}^{initial} \times \frac{e}{E_t}
\end{align}
This synergistic combination—using an external teacher for rapid, stable initialization, followed by internal self-distillation for generalized regularization—forms the core mechanism that allows ASKD-Whisper to surpass standard compression bounds. The detailed pseudo-code for ASKD is provided in Algorithm 1. (In our experiments, we set $\alpha_{\text{AKD}}^{initial}=1$, $\alpha_{\text{SKD}}^{initial}=0.8$, $\lambda=0.5$, $E_\text{t}=10$, and $E_\text{w}=2$).

\begin{algorithm}
\footnotesize
\label{algo:askd}
\caption{Adaptive Self-knowledge Distillation}
\begin{algorithmic}[1]
\Require Audio data $X_{\text{audio}}$, Text token $Y_{\text{token}}$, Student model $\theta_S$, Teacher model $\theta_T$, Pre-trained audio extractor $\theta_{E}$, Cross-entropy losses $\mathcal{L}_{\text{CE}}$, Temperature $\tau$, AKD alpha $\alpha_{\text{AKD}}$, SKD alpha $\alpha_{\text{SKD}}$, Threshold $\lambda$, Warm up Epoch $E_w$, Total Epoch $E_t$, Softmax $\sigma$, learning rate $\eta$

\State Initialize student decoder parameters $\theta_S$
\State Freeze $\theta_{E}$
\For{$\{e = 0, 1, \dots, E_t\}$}
    \State $\mathbf{x}, \mathbf{y} \xleftarrow{} MiniBatchSampler(X_{\text{audio}},Y_{\text{token}})$
    \State $\mathbf{x}_{a} \xleftarrow{} \theta_{E}(x)$
    \If{$\alpha_{\text{AKD}}^{e} > \lambda$}
        \State $\mathcal{L}_{S} \xleftarrow{} \mathcal{L}_{\text{CE}}(\theta_{S}(\mathbf{y},\mathbf{x}_{a}),\mathbf{y})$
        \State $P_S, P_T \xleftarrow{} \sigma(\theta_i(\mathbf{y},\mathbf{x}_{a}) / \tau), \forall i \in \{S,T\}$
        \State Form $\mathcal{L}_{\text{AKD}}$ as given by Eq. (\ref{eq:AKD_loss})
        \State $\mathcal{L}_{\text{total}} = \mathcal{L}_{S} + \mathcal{L}_{\text{AKD}}$
    \Else
        \State $\theta_T \xleftarrow{} \theta_{S}^{(e-1)}$
        \State $P_S,P_T \xleftarrow{} \sigma(\theta_i(\mathbf{y},\mathbf{x}_{a})), \forall i \in \{S,T\}$
        \State Form $\mathcal{L}_{\text{SKD}}$ as given by Eq. (\ref{eq:SKD_loss})
        \State $\mathcal{L}_{\text{total}} = \mathcal{L}_{\text{SKD}}$
    \EndIf
    \State Update $\theta_S \xrightarrow{}{} \theta_S - \eta \nabla \mathcal{L}_{\text{total}}$
    \If{$e < E_w$}        
        \State $\alpha_{\text{AKD}}^{e} \gets \alpha_{\text{AKD}}^{initial}$    
    \Else
        \If{$\alpha_{\text{AKD}}^{e} > \lambda$}
            \State Update $\alpha_{\text{AKD}}^{e}$ using Eq. (\ref{eq:AKD_update})
        \Else
            \State Update $\alpha_{\text{SKD}}^{e}$ using Eq. (\ref{eq:SKD_update})
        \EndIf
    \EndIf
\EndFor
\end{algorithmic}
\end{algorithm}

\section{Experiment Results}

\subsection{Data Description}
To comprehensively evaluate the robustness of our approach, we curated a highly diverse training corpus totaling 1,620 hours. This comprises 960 hours of LibriSpeech \cite{Panayotov_15} (clean audiobook narrations), 453 hours of TED-LIUM release3 \cite{Hernandez_18} (academic/political conference speeches), 24 hours of LJSpeech \cite{Keith_17}, 105 hours of Earnings-22 \cite{Del_22} (real-world financial meetings laden with technical jargon), and 78 hours of the AMI Meeting Corpus \cite{Carletta_17} (spontaneous, overlapping conversational speech).

To rigorously test the out-of-distribution (OOD) generalization capacity of our model, we additionally evaluated on two completely unseen datasets: GigaSpeech \cite{Chen_21} (multi-domain corpus spanning podcasts and YouTube) and VoxPopuli \cite{Wang_21} (European Parliament recordings featuring heavy non-native English accents).

\subsection{Model Architecture}
To maximize data efficiency, ASKD-Whisper leverages the pre-trained, frozen Whisper encoder \cite{Radford_23}, which inherently contains rich acoustic representations learned from 680,000 hours of audio. The primary architectural innovation lies in the decoder. We replaced the standard deep multi-head attention blocks with a custom lightweight Transformer decoder utilizing SwiGLU \cite{Shazeer_20} activations, which significantly enhances parameter efficiency and non-linear expressivity.

We implemented two variants:
\begin{itemize}[leftmargin=*]
    \item \textbf{ASKD-Whisper-small} (152M parameters): Designed for ultra-low latency edge deployment. It pairs the Whisper-small encoder with a 3-layer custom SwiGLU decoder.
    \item \textbf{ASKD-Whisper-large} (740M parameters): Designed to achieve SOTA accuracy while maintaining rapid inference. It pairs the Whisper-large-v3 encoder with a highly compressed 3-layer custom decoder.
\end{itemize}

\subsection{Efficacy of the Distillation Paradigms}
\label{sec:KD_methods}

To isolate the impact of our proposed ASKD framework, we conducted an ablation study tracking Word Error Rate (WER) 
\begin{table}[!h]
    \centering
    \caption{Comparison of WER across knowledge distillation methods. The best method is in red boldface and the second best is blue.}
    \label{tab:KD_compare}
    \renewcommand{\arraystretch}{1.2}
    \resizebox{\columnwidth}{!}{ 
    \begin{tabular}{c c c c c}
        \toprule
        \multirow{2}{*}{\textbf{Method}} & \multirow{2}{*}{\textbf{Model}} & \multicolumn{3}{c}{\textbf{Dataset}} \\ 
        \cmidrule(lr){3-5}
        & & {Test-clean} & {Test-other} & {Earnings-22} \\ 
        \midrule
        Baseline (Teacher) & Whisper-small & \textcolor{blue}{3.05} & \textcolor{red}{\textbf{7.25}} & 13.0 \\
        Standard KD + PL & distill-Whisper-S & 3.45 & 7.71 & 13.2 \\
        SKD Only & ASKD-Whisper-small & 5.82 & 11.5 & 16.8 \\
        \textbf{AKD (ours)} & ASKD-Whisper-small & 3.06 & 7.66 & \textcolor{blue}{12.2} \\
        \textbf{ASKD (ours)} & ASKD-Whisper-small & \textcolor{red}{\textbf{2.95}} & \textcolor{blue}{7.63} & \textcolor{red}{\textbf{11.8}} \\
        \bottomrule
    \end{tabular}
    }
\end{table}
\begin{table*}[ht]
    \caption{Comparison of WER scores between ASKD-Whisper and state-of-the-art (SOTA) models. ASKD-Whisper achieves competitive or superior results despite utilizing a fraction of the training data and parameters.}
    \label{tab:SOTA_WER}
    \renewcommand{\arraystretch}{1.3}
    \centering
    \resizebox{2\columnwidth}{!}{
    \begin{tabular}{c c c c c c c c c}
        \toprule
        \multirow{2}{*}{\textbf{Model}} & \multirow{2}{*}{\textbf{Training Data (h)}} & \multirow{2}{*}{\textbf{Params (B)}} & \multicolumn{6}{c}{\textbf{Dataset (WER $\downarrow$)}} \\ 
        \cmidrule(lr){4-9}
        & & & {Test-clean} & {Test-other} & {TED-LIUM} & {AMI} & {Earnings-22} & {Average} \\
        \midrule
        Whisper-large-v3 \cite{Radford_23} & $680,000$ & $1.50$ & $\textcolor{blue}{2.01}$ & $\textcolor{blue}{3.91}$ & $3.88$ & $16.0$ & $\textcolor{blue}{11.4}$ & $7.44$\\ 
        distil-Whisper-L \cite{Gandhi_23} & $22,000$ & $\textcolor{blue}{0.75}$& $2.54$ & $5.19$ & $3.86$ & $15.1$ & $11.8$ & $7.70$\\
        Canary \cite{Krishna_24} & $31,200$ & $1.00$& $\textcolor{red}{\mathbf{1.48}}$ & $\textcolor{red}{\mathbf{2.93}}$ & $\textcolor{blue}{3.56}$ & $13.9$ & $12.2$ & $6.82$\\
        CrisperWhisper \cite{Zusag_24} & $\textcolor{blue}{10,000}$ & $1.50$& $\textcolor{blue}{1.82}$ & $4.00$ & $\textcolor{red}{\mathbf{3.20}}$ & $\textcolor{red}{\mathbf{9.89}}$ & $12.9$ & $\textcolor{red}{\mathbf{6.36}}$\\
        \textbf{ASKD-Whisper-large (ours)} & $\textcolor{red}{\mathbf{1,634}}$ & $\textcolor{red}{\mathbf{0.74}}$ & $2.34$ & $4.52$ & $3.89$ & $\textcolor{blue}{10.6}$ & $\textcolor{red}{\mathbf{10.5}}$ & $\textcolor{blue}{6.37}$ \\
        \bottomrule
    \end{tabular}
    }
\end{table*} 
across different KD methodologies using the ASKD-Whisper-small architecture. The results are presented in Table \ref{tab:KD_compare}.

The empirical evidence strongly supports our hypothesis. Standard KD paired with Pseudo-Labeling (PL) performs worse than the teacher across all domains, illustrating the "capacity gap" limitation where the student fails to map complex teacher logits to limited parameters. Conversely, our dynamic AKD module actively prevents this over-reliance, yielding a 1.0\% absolute improvement on the challenging Earnings-22 dataset compared to Standard KD. The full ASKD framework—integrating AKD with SKD regularization—achieves the best results, actually surpassing the massive teacher model on Test-Clean (2.95\%) and Earnings-22 (11.8\%).

\subsection{Quantitative State-of-the-Art Comparison}
\label{sec:SOTA_compare}
We benchmarked ASKD-Whisper-large against current State-of-the-Art (SOTA) models, including the original Whisper-large-v3, distill-Whisper-large-v3, Canary, and CrisperWhisper. 

As detailed in Table \ref{tab:SOTA_WER}, ASKD-Whisper-large significantly outperforms distill-Whisper-large-v3 across the board (average 6.37\% vs 7.70\%), despite distill-Whisper utilizing $13\times$ more training data. The strength of ASKD is particularly evident in highly complex, noisy environments: on the AMI (overlapping conversational speech) and Earnings-22 (specialized financial jargon) datasets, ASKD-Whisper-large reduces the WER by a massive 5.4\% and 0.9\% absolute, respectively, compared to its teacher, Whisper-large-v3. This confirms that ASKD successfully regularizes the model against the specific vulnerabilities and hallucinations prevalent in the teacher's distribution. To ensure statistical rigor, all evaluations report the mean WER over 3 random seeds. Seed variation was minimal ($\le 0.05\%$ WER), confirming the stability of the ASKD framework.

\subsection{Latency and Out-of-Domain Generalization}
The primary objective of model compression is accelerating inference speed without sacrificing robustness. Table \ref{tab:Rel_latency} presents the relative decoding latency and generalization performance on completely unseen distributions (VoxPopuli and GigaSpeech). To transparently evaluate efficient offline decoding, we measured latency using the Real-Time Factor (RTF). Measurements were conducted with a batch size of 1 using deterministic greedy decoding (temperature = 0.0) on a single NVIDIA A100 80GB GPU.

\begin{table}[h]
    \centering
    \caption{Comparison of relative latency and out-of-domain (OOD) generalization performance.}
    \label{tab:Rel_latency}
    \renewcommand{\arraystretch}{1.2}
    \resizebox{\columnwidth}{!}{ 
    \begin{tabular}{c c c c}
        \toprule
        \multirow{2}{*}{\textbf{Model}} & \multirow{2}{*}{\textbf{Relative Latency}} & \multicolumn{2}{c}{\textbf{Dataset (OOD)}} \\ \cmidrule(lr){3-4}
        & & {Voxpopuli} & {Gigaspeech}\\ 
        \midrule
        Whisper-large-v3 & $\times 1.0$ ($659\, \text{ms}$) & 9.54 & 10.0 \\ 
        \textbf{ASKD-Whisper-large} & $\times$ \textbf{5.0} ($132\, \text{ms}$) & 9.59 & 10.2 \\
        \bottomrule
    \end{tabular}
    }
\end{table}

While drastically reducing the deep decoder stack into a shallow 3-layer SwiGLU architecture natively provides a significant $5\times$ speedup during auto-regressive generation, such aggressive architectural truncation typically leads to severe performance degradation. Crucially, because it is trained via the ASKD framework, ASKD-Whisper is able to maintain this high inference speed while exhibiting a negligible WER degradation of less than 0.2\% compared to the 1.5B parameter teacher on OOD datasets featuring heavy non-native accents (VoxPopuli) and diverse podcast environments (GigaSpeech).

\subsection{Information-Theoretic Analysis of the Distillation Threshold}
\begin{figure}[h]
    \centering
    \includegraphics[width=0.9\columnwidth]{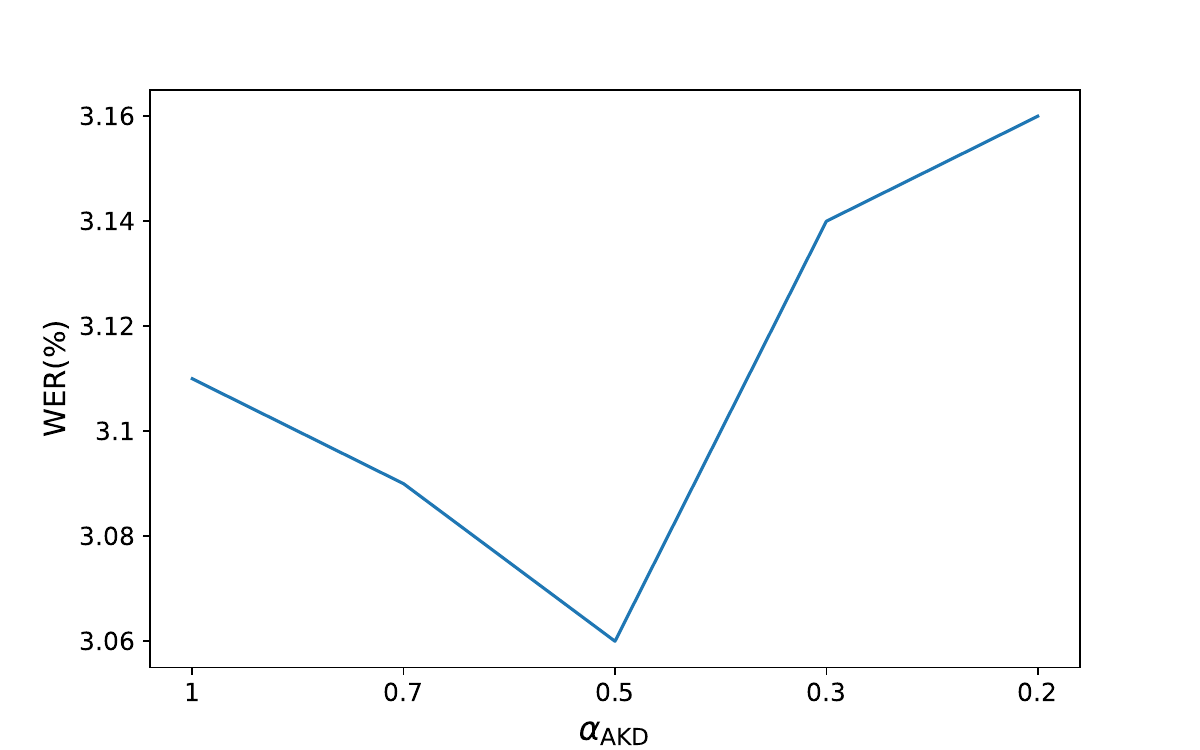}
    \caption{Impact of the minimal $\alpha_{\text{AKD}}$ threshold on final WER. A threshold of 0.5 provides the optimal balance between teacher initialization and self-regularization.}
    \label{fig:alpha_wer}
\end{figure}

A critical hyperparameter in our framework is the minimal threshold $\lambda$, which dictates when the model transitions from AKD to SKD. Fig. \ref{fig:alpha_wer} illustrates the effect of varying this minimal threshold on the Test-Clean dataset.

From a representation learning perspective, enforcing a high threshold (e.g., 0.9) maintains a strict dependency on the teacher, leading to semantic saturation where the student memorizes noise. Conversely, transitioning to SKD too early (e.g., threshold 0.1) deprives the student of essential foundational mapping. The optimal point at $\alpha_{\text{AKD}} = 0.5$ confirms that a balanced curriculum—spending equal phase energy on external acquisition and internal regularization—yields the optimal representation geometry for ASR tasks. Furthermore, ASKD is highly robust to this hyperparameter: our sensitivity analysis revealed that varying the threshold to $\lambda \in \{0.3, 0.7\}$ resulted in marginal WER variations of $\le 0.05\%$, justifying the stability of our explicit phase transition.

\section{Conclusion}
In this study, we addressed the fundamental capacity gap and over-reliance problems inherent in standard knowledge distillation for speech models. We introduced Adaptive Self-Knowledge Distillation (ASKD), a dynamic paradigm that strategically shifts from teacher-guided learning to self-regularized training. Applied to the Whisper architecture, our ASKD-Whisper models demonstrated that strategic decoupling from a teacher model actually allows a student to surpass the teacher's limitations in complex, noisy environments. ASKD-Whisper achieved a $5\times$ acceleration in inference latency while establishing a 1.07\% lower WER on specialized datasets compared to its 1.5B parameter teacher. Future research will explore scaling the ASKD framework to cross-lingual distillation to validate its efficacy on low-resource languages.

\section*{Limitations}
While ASKD-Whisper demonstrates significant improvements in inference latency and out-of-distribution generalization, it has notable limitations. First, our empirical validation is entirely restricted to English-language corpora. The effectiveness of the dynamic ASKD threshold ($\lambda$) and the SwiGLU decoder's representational capacity on morphologically complex or low-resource languages remains unverified. Second, our evaluations primarily target single-utterance or short-form speech segments. For long-form audio transcription (e.g., multi-hour podcasts or meetings), sequence-to-sequence models inherently suffer from text repetition and severe hallucinations. Whether the label-smoothing regularization effect of the SKD phase scales effectively to suppress insertion errors during long-form sequential decoding requires dedicated future investigation.

\section*{Ethical Considerations}
The deployment of ASKD-Whisper contributes positively to the sustainable development of AI. By achieving a 5$\times$ speedup in auto-regressive decoding, the architecture significantly reduces the energy consumption and carbon emissions associated with large-scale ASR cloud deployments. However, as with all models trained on internet-sourced corpora, inherent demographic and socio-economic biases may exist in the acoustic representations. We advocate for rigorous fairness evaluations before deploying these systems in critical decision-making contexts.

\bibliography{custom}

\end{document}